\documentclass[10pt,twocolumn,letterpaper]{article}

\usepackage{cvpr}              %

\usepackage{graphicx}
\usepackage{amsmath}
\usepackage{amssymb}
\usepackage{booktabs}
\usepackage{microtype}
\usepackage{xcolor}
\usepackage{array}   %
\usepackage{pifont}       %
\usepackage{bbding}       %
\usepackage{fontawesome}  %
\usepackage{tcolorbox}
\usepackage{wrapfig}
\usepackage{colortbl}
\usepackage{graphicx}
\usepackage{multirow}
\usepackage{makecell}
\usepackage{enumitem} %
\usepackage{soul}
\usepackage{tabularx}

\newcommand{\blfootnote}[1]{%
  \begingroup
  \renewcommand\thefootnote{}%
  \footnote{#1}%
  \addtocounter{footnote}{-1}%
  \endgroup
}

\renewcommand{\paragraph}[1]{\vspace{.5em}\noindent\textbf{#1}}

\usepackage{xspace}

\colorlet{colorFst}{Green!25}       %
\colorlet{colorSnd}{SpringGreen!45} %
\colorlet{colorTrd}{Yellow!30}      %
\colorlet{colorLow}{darkgray!30}    %
\newcommand{\fs}{\cellcolor{colorFst}\bf}   %
\newcommand{\nd}{\cellcolor{colorSnd}}      %
\newcommand{\rd}{\cellcolor{colorTrd}}      %

\newcommand{\cmark}{\ding{51}}%

\newcommand{\authorskip}{\hspace{2em}}

\definecolor{cvprblue}{rgb}{0.21,0.49,0.74}
\usepackage[pagebackref,breaklinks,colorlinks,allcolors=cvprblue]{hyperref}

\def\paperID{5} %
\def\confName{CVPR}
\def\confYear{2026}

\title{Spatial-aware Vision Language Model for Autonomous Driving}

\author{
 Weijie Wei$^{1,2,*}$
 \authorskip Zhipeng Luo$^{1,\dagger}$
 \authorskip Ling Feng$^{1}$
 \authorskip Venice Erin Liong$^{1,\ddagger}$\\
 $^1$Motional \qquad $^2$University of Amsterdam\\
}

\definecolor{turquoise}{cmyk}{0.65,0,0.1,0.3}
\definecolor{purple}{rgb}{0.65,0,0.65}
\definecolor{dark_green}{rgb}{0, 0.5, 0}
\definecolor{orange}{rgb}{0.8, 0.6, 0.2}
\definecolor{red}{rgb}{0.8, 0.2, 0.2}
\definecolor{darkred}{rgb}{0.6, 0.1, 0.05}
\definecolor{blueish}{rgb}{0.0, 0.3, .6}
\definecolor{light_gray}{rgb}{0.7, 0.7, .7}
\definecolor{pink}{rgb}{1, 0, 1}
\definecolor{greyblue}{rgb}{0.25, 0.25, 1}

\newcommand{\Ours}{LVLDrive\xspace}  %
\newcommand{\OurFusion}{Gradual Fusion Q-Former\xspace}
\newcommand{\OurDataset}{SA-QA\xspace}

\begin{document}
\maketitle

\begin{abstract}
While Vision-Language Models (VLMs) show significant promise for end-to-end autonomous driving by leveraging the common sense embedded in language models, their reliance on 2D image cues for complex scene understanding and decision-making presents a critical bottleneck for safety and reliability. Current image-based methods struggle with accurate metric spatial reasoning and geometric inference, leading to unreliable driving policies. To bridge this gap, we propose LVLDrive (LiDAR-Vision-Language), a novel framework specifically designed to equip existing VLMs with robust 3D metric spatial understanding for autonomous driving by incorporating LiDAR point clouds as an extra input modality. A key challenge lies in mitigating the catastrophic disturbance introduced by disparate 3D data to the pre-trained VLMs. To this end, we introduce a Gradual Fusion Q-Former that incrementally injects LiDAR features, ensuring the stability and preservation of the VLM's existing knowledge base. Furthermore, we develop a spatial-aware question–answering (SA-QA) dataset to explicitly teach the model advanced 3D perception and reasoning capabilities. Extensive experiments on driving benchmarks demonstrate that LVLDrive achieves superior performance compared to vision-only counterparts across scene understanding, metric spatial perception, and reliable driving decision-making. Our work highlights the necessity of explicit 3D metric data for building trustworthy VLM-based autonomous systems.
\end{abstract}    
\blfootnote{$^*$Work done as an intern at Motional. $^\dagger$Project lead. $^\ddagger$Corresponding author.}
\section{Introduction}
\label{sec:intro}

\begin{figure}[t] \centering
    \includegraphics[width=0.9\linewidth]{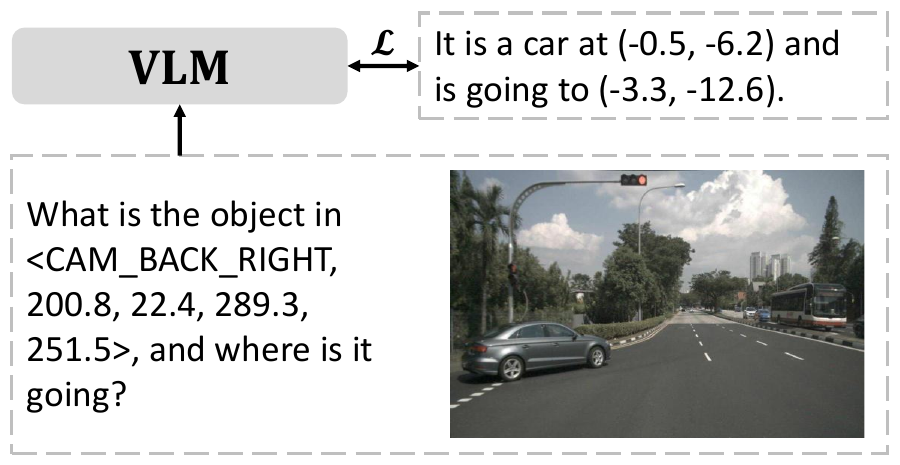}
    \\
    \makebox[0.98\linewidth]{(a) Image-based VLM for autonomous driving.}
    \\ \vspace{0.5em}
    \includegraphics[width=0.9\linewidth]{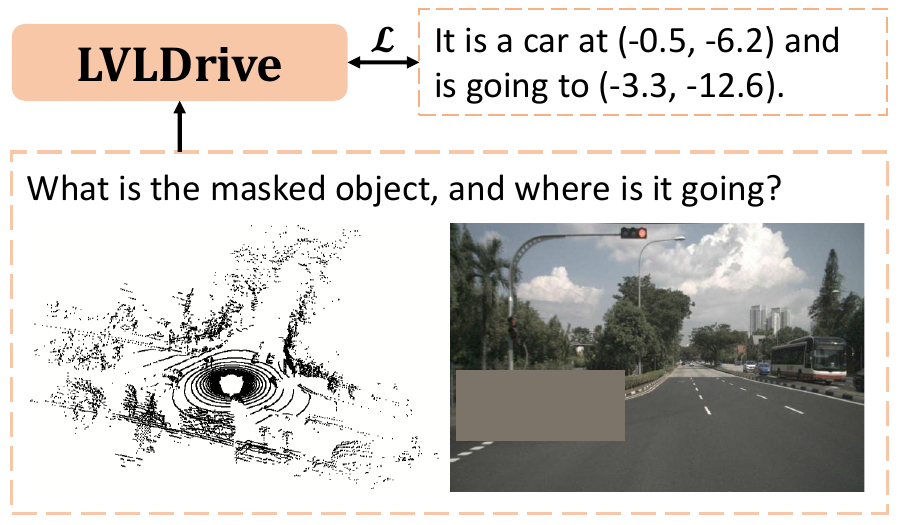}
    \\
    \makebox[0.98\linewidth]{(b) \Ours (Ours).}
    \caption{Typical image-based VLMs \textbf{(a)} take only images as input and train the LLM through image-conditioned question answering. In contrast, \Ours \textbf{(b)} leverages both image and LiDAR inputs and incorporates carefully designed spatial-aware QAs to encourage effective LiDAR integration and enhance spatial understanding.}
    \label{fig:teaser}
    \vspace{-1em}
\end{figure}

Large vision–language models (VLMs) have recently emerged as a promising foundation for end-to-end autonomous driving systems. By coupling visual perception with language-native reasoning and explanation, VLM-based agents can describe scenes, justify decisions, and expose intermediate rationales that were opaque in earlier pipelines \cite{cSima_2024_DriveLM, shao2024lmdrive, pan2024vlp, li2024llada}. However, despite rapid progress, a critical limitation persists: contemporary VLMs predominantly consume 2D imagery and thus struggle with metric spatial understanding—precise distances, extents, occlusions, and interactions in 3D. These capabilities are essential for safe planning in dense, multi-agent urban scenes, especially in autonomous driving. Recent benchmarks that isolate spatial queries corroborate this gap, showing that strong general-purpose VLMs underperform on driving-critical spatial reasoning tasks \cite{kTian_2025_NuScenes-SpatialQA, yang2025thinking}.

Why is this challenging? Inferring geometry from monocular or multi-view images is ill-posed under occlusions, adverse weather, and viewpoint changes. Even when bird’s-eye-view (BEV) features or video context are introduced, image-only models tend to confuse appearance with metric structure, leading to brittle estimates of range, free space, and collision risk. As a result, VLM-based driving agents may produce fluent rationales yet fail at the quantitative spatial inferences that underwrite reliable control.
LiDAR offers a complementary pathway. Its calibrated 3D point measurements provide direct, long-range, and lighting-invariant geometric cues for scene layout, drivable area, and inter-agent relations. Prior works have demonstrated that integrating LiDAR with images can significantly enhance the perception performance in autonomous driving scenes \cite{liang2022bevfusion,wang2025mv2dfusion}. While integrating LiDAR with VLMs seems like a natural next step, it remains challenging. Pre-trained VLMs are optimized on massive image–text corpora but not on LiDAR–image–text multimodal data. Consequently, naively injecting a disparate 3D representation can severely disrupt their learned alignments, resulting in suboptimal linguistic competence and visual grounding.

To this end, we introduce LVLDrive (LiDAR, Vision and Language), a framework that equips VLMs with robust 3D metric spatial understanding while preserving their learned knowledge (see~\cref{fig:teaser}).
The key component is the \OurFusion that incrementally injects LiDAR embeddings into the VLM via a gated attention mechanism.
Instead of exposing the model to unaligned 3D signals all at once, LVLDrive gradually introduces LiDAR features during training, allowing cross-modal attention to adapt without drifting from the original visual–linguistic manifold. The learnable gate determines when and to what extent to rely on LiDAR, enabling graceful fallback to image cues when 3D observations are sparse or noisy.
To explicitly promote spatial reasoning and LiDAR integration, we construct the \OurDataset dataset, a spatial-aware visual question–answering (QA) dataset derived from ground-truth 3D annotations in standard driving scenes.
The dataset focuses on metric and relational QA pairs that are critical to autonomous driving, pairing linguistically natural questions with unambiguous, 3D-grounded answers and complementing prior instruction-tuned driving corpora \cite{cSima_2024_DriveLM, ding2024holisticnuinstruct, wang2025omnidrive}.
A subset of QA pairs further incorporates modality masking, forcing the VLM to rely on LiDAR and strengthening cross-modal alignment.

We evaluate LVLDrive on recent language–based driving benchmarks emphasizing spatial reasoning and decision reliability.
Across scene understanding, metric spatial perception, and planning-relevant intent prediction, LVLDrive consistently outperforms vision-only VLM baselines, narrowing the gap between the commonsense strengths of VLMs and the metric precision demanded by safety-critical driving. By integrating LiDAR in a stability-preserving manner and supervising with spatial-aware language tasks, our approach improves the reliability of LLM-based autonomous systems and underscores the necessity of explicit 3D metric cues for trustworthy end-to-end driving. The contributions of this work are threefold:
(1) We introduce \Ours, augmenting a pre-trained VLM with a \OurFusion to incrementally inject LiDAR embeddings while preserving visual–linguistic priors;
(2) We construct a spatial-aware QA dataset (\OurDataset) on top of the nuScenes dataset and its ground-truth annotations, to boost the spatial understanding and reasoning of \Ours;
(3) We conduct extensive experiments and ablation studies to quantify the performance gains and to clarify how LiDAR injection and fine-tuning on our proposed spatial-aware dataset contribute to faithful 3D spatial reasoning.
\section{Related Work}
\label{sec:related_work}

\paragraph{Language–Driving Datasets.}
Early work grounded natural-language commands and explanations in real driving scenes, \eg Talk2Car~\cite{deruyttere2019talk2car} for referred-object commands, the BDD-X~\cite{kim2018bddx} explanation corpus, and BDD-OIA~\cite{xu2020bdd-oia} for object-induced action with textual rationales.
Recently, QA datasets have scaled substantially.
nuScenes-QA~\cite{qian2024nuscenesqa} standardized multi-view urban driving VQA, while LingoQA~\cite{marcu2024lingoqa} extended short driving videos with human truthfulness evaluation.
DriveLM \cite{cSima_2024_DriveLM} introduced graph-structured chain-of-thought reasoning and released DriveLM-Data on nuScenes and CARLA, along with a GVQA protocol spanning perception, prediction, and planning.
BEV-centric benchmarks further emphasized global scene reasoning, including Talk2BEV-Bench \cite{choudhary2024talk2bev}, BEV-TSR \cite{tang2025bevtsr} for text-to-scene retrieval, and ChatBEV-QA \cite{xu2025chatbev}.
Instruction-style supervision also appears in datasets such as NuInstruct \cite{ding2024holisticnuinstruct}, paired with BEV-injected MLLMs.
OmniDrive~\cite{wang2025omnidrive} proposed a holistic language-driving dataset with counterfactual reasoning.
Most prior datasets emphasize general scene understanding and decision-making. In contrast, nuScenes-SpatialQA \cite{kTian_2025_NuScenes-SpatialQA} specifically benchmarks spatial reasoning in VLMs.
While our SA-QA dataset shares this motivation, it focuses more on planning-oriented spatial reasoning and cross-modal interaction between LiDAR and vision.

\begin{figure*}[t] 
\centering
    \includegraphics[width=0.98\textwidth]{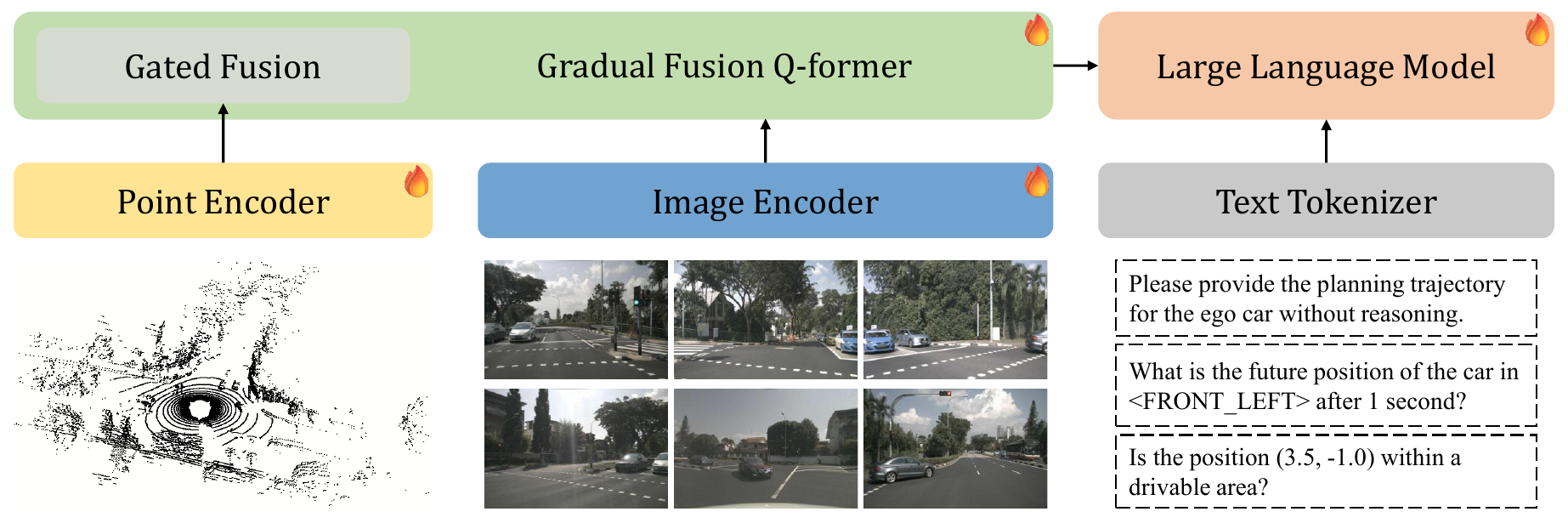}
    \caption{\textbf{Overview of \Ours.} \Ours takes text, images, and point clouds as multimodal inputs and employs three pretrained encoders, an LLM, and a \OurFusion module that bridges visual and linguistic representations to generate task-specific responses. The fire symbol denotes trainable components.
    }
    \label{fig:method_overview}
    \vspace{-1em}
\end{figure*}

\paragraph{VLMs for Autonomous Driving.}
Vision-Language Models have been widely adopted in autonomous driving, primarily relying on image inputs for human-interpretable scene understanding and trajectory planning.
DriveLM~\cite{cSima_2024_DriveLM} introduced a VLM-based agent for graph VQA and end-to-end control; LMDrive~\cite{shao2024lmdrive} integrated LLM reasoning with closed-loop control; VLP~\cite{pan2024vlp} bridged perception, text, and planning with language supervision; and LLaDA~\cite{li2024llada} adapted policies to new domains using LLM priors. GPT-style planners such as GPT-Driver~\cite{mao2023gptdriver} and DriveGPT4~\cite{xu2024drivegpt4} explored autoregressive action generation with rationales.
DriveVLM~\cite{tian2024drivevlm} and DriVLMe~\cite{huang2024drivlme} developed multimodal driving agents for QA and navigation.
BEV-aware multimodal models like BEV-InMLLM~\cite{ding2024holisticnuinstruct} improved global spatial context, while V2X-VLM~\cite{you2024v2x} leveraged infrastructure views. More recently, VLA families such as OpenDriveVLA~\cite{zhou2025opendrivevla} and AutoVLA~\cite{zhou2025autovla} unified perception, reasoning, and action in a single autoregressive policy incorporating feasibility and physics constraints; InternDrive~\cite{zhang2024interndrive} focused on scenario understanding with MLLMs. Unlike these image-based approaches, LiDAR-LLM~\cite{yang2025lidarllm} pioneers the use of LLMs for driving-scene understanding directly from LiDAR inputs, aiming to extend VLMs’ comprehension to more challenging real-world 3D scenes. In contrast, our work explores the complementary strengths of both image and LiDAR within a unified framework, leveraging the strong reasoning capabilities of pretrained VLMs together with the 3D metric information from LiDAR point clouds.

\paragraph{Spatial Understanding in VLMs.}
In recent years, generic VLMs~\cite{hurst2024gpt, bai2025qwen2, liu2023visualllava, wu2024deepseek} have achieved remarkable progress, exhibiting strong reasoning and problem-solving abilities in open-ended tasks. However, several studies~\cite{jia2025OmniSpatial, ma20253dsrbench, liu2025SpatialMQA} have evaluated the spatial understanding and reasoning capabilities of VLMs and found they are far from satisfactory—even with enhancements such as chain-of-thought prompting~\cite{yang2025thinking}. To address these gaps, several recent studies on the understanding of indoor scenes-including SpatialVLM~\cite{bChen_2024_SpatialVLM} and SpatialRGPT~\cite{acCheng_2024_SpatialRGPT}—have explored fine-tuning strategies with spatially enriched task-specific datasets. In autonomous driving contexts, where accurate spatial reasoning is essential for planning and safety, benchmarking results on NuScenes-SpatialQA~\cite{kTian_2025_NuScenes-SpatialQA} similarly reveal that current VLMs still fall short. These findings highlight the need to explore more capable approaches to effectively enhance spatial understanding and reasoning in autonomous driving scenarios.

\section{Methodology} \label{sec:method}

\subsection{Preliminaries} \label{sec:preliminaries}
\paragraph{Vision-Language Model (VLM).}
A VLM typically comprises three main components:
a language model pretrained on large-scale text corpora using next-token prediction objectives~\cite{touvron2023llama, ouyang2022traininginstructgpt};
a vision encoder pretrained on visual tasks through self-supervised learning~\cite{kHe_2021_MAE, xChen_2021_MoCoV3, mCaron_2021_DINO}, contrastive learning~\cite{aRadford_2021_CLIP, jLi_2022_BLIP} or supervised learning~\cite{aDosovitskiy_2021_ViT, he2017maskrcnn, xie2021segformer};
and a projector that injects visual features into the language model.
The projector can be a simple MLP, as in LLaVA~\cite{liu2023visualllava, hLiu_2024_LLaVA1.5}, or a more structured Q-Former~\cite{jLi_2023_BLIP2} built with cross-attention layers.
In the Q-Former, learnable queries attend to external visual features, enabling the model to retrieve and compress relevant information into a fixed length of latent tokens.
Since point cloud encoders yield variable-length outputs due to sparsity, we adopt a Q-Former-style projector to produce fixed-length token outputs, enabling robust interfacing with the language model.

\begin{figure}[t] \centering
    \includegraphics[width=0.9\linewidth]{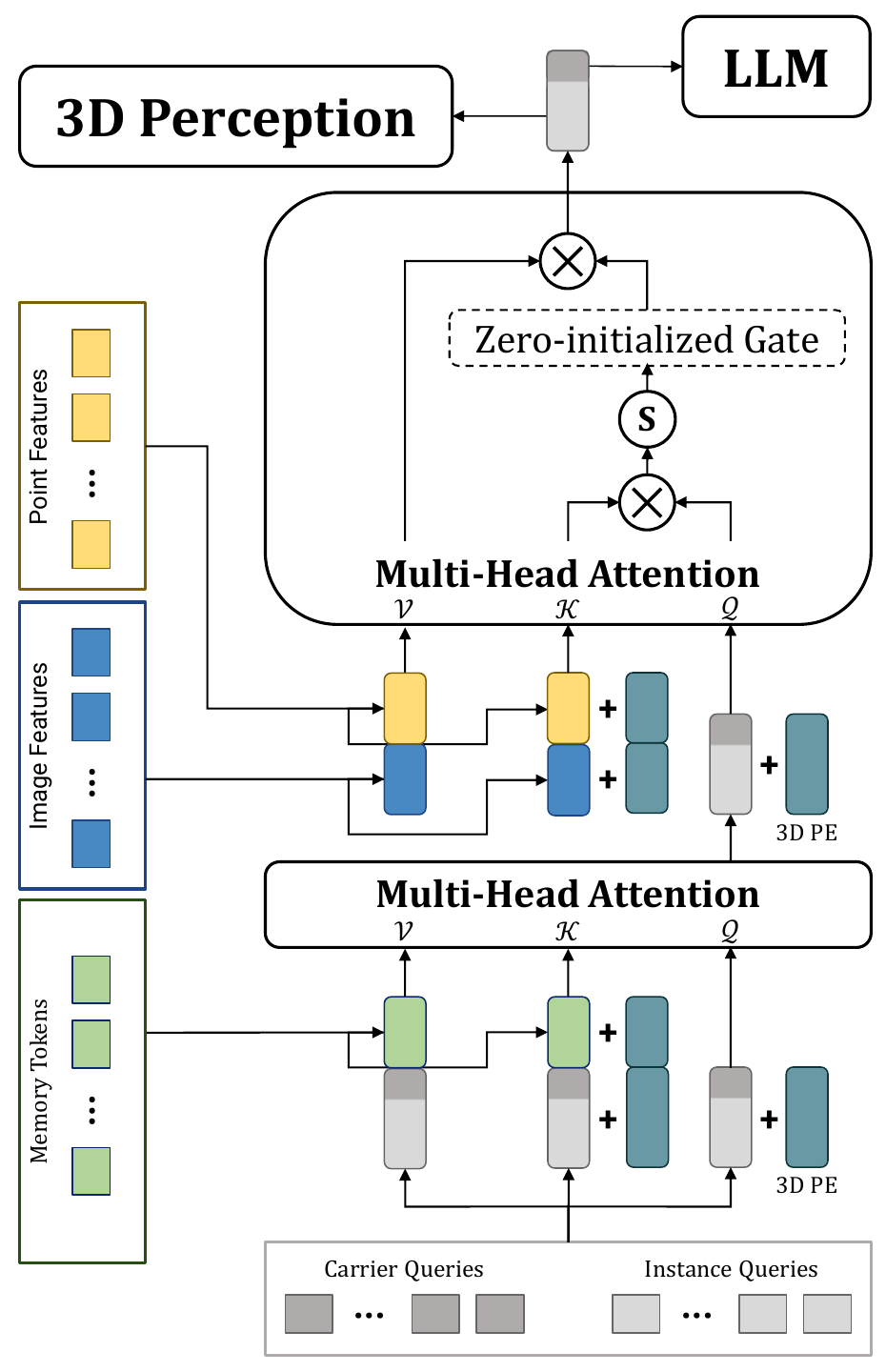}
    \caption{\textbf{\OurFusion}. Each block contains two multi-head attention layers. The first layer uses learnable carrier and instance tokens as queries, keys, and values; the keys and values are extended with memory tokens when the memory bank is non-empty, and both queries and keys are augmented with 3D positional embeddings (3D PE) derived from reference points. The second layer introduces image and point features as keys and values, with queries and keys again augmented using their corresponding 3D positional embeddings. Note that the zero-initialized gate is applied only to point features. At the end of the module, the output queries are supervised by 3D perception objectives and LLM gradients separately.}
    \label{fig:fusion_module}
    \vspace{-1em}
\end{figure}

\paragraph{Q-Former 3D Block.}
Q-Former 3D block is an upgraded version of the Q-Former and was proposed in OmniDrive~\cite{wang2025omnidrive}.
While it shares the core idea and architecture of a standard Q-Former—\ie two multi-head attention (MHA) layers per block stacked in depth—OmniDrive modifies the input formulation so that the architecture can decode object-centric features from images.
As illustrated in~\cref{fig:fusion_module}, the first MHA layer in each block uses a shared set of learnable instance tokens as queries, keys, and values. Each query corresponds to a reference point in 3D space; thus, 3D positional embeddings derived from these reference points are added to both the queries and keys. When a memory bank is available (carrying information from previous frames), memory tokens are concatenated to the keys and values to incorporate temporal context.
The second MHA layer takes the output of the first layer as queries and the patch-wise image embeddings as keys and values, enabling retrieval and compression of visual features. Positional embeddings from the reference points are again added to the queries and keys to maintain spatial correspondence.
Building on this design, OmniDrive further concatenates carrier tokens with the instance tokens and processes them through multiple Q-Former 3D blocks. The carrier tokens are then passed to the LLM, while the instance tokens are supervised with 3D perception objectives.
Our fusion module is constructed on top of this Q-Former 3D block introduced by OmniDrive.

\subsection{Framework Overview} \label{sec:method_overview}
As illustrated in~\cref{fig:method_overview}, \Ours takes text, images, and point clouds as multimodal inputs to generate task-specific responses via an LLM.
The overall architecture comprises three pretrained encoders (for text, image, and point cloud), an LLM, and a \OurFusion that serves as a projector between visual and linguistic representations.
The text tokenizer, image encoder, LLM, and \OurFusion—implemented with stacked multi-head attention blocks—have been jointly pretrained without point cloud input, enabling well-aligned feature representations across text-image modalities.
This strong alignment, however, does not extend to the point cloud encoder.
Unlike the image modality, large-scale LiDAR-text data for joint pretraining is scarce. This lack of alignment means that naively injecting point cloud features into the VLM could disrupt the well-learned text-image representations and lead to suboptimal performance. Our core challenge, therefore, is to effectively integrate these ``out-of-domain" features under limited data conditions.
To address this, we introduce a unified \OurFusion that progressively integrates the newly acquired point cloud features into the carrier tokens and jointly retrieves multimodal representations for downstream reasoning.

\subsection{\OurFusion} \label{sec:fusion_module}
As discussed in~\cref{sec:method_overview}, a key challenge lies in how to inject point cloud features into an existing VLM where text and image representations are already well aligned. To address this, we investigate the Q-Former 3D block and identify its three inputs:
1) a set of learnable query tokens,
2) tokens from the memory bank, which encode object-centric features from historical frames, and
3) image feature tokens, through which information from the current frame is incorporated.
Each token is associated with either a specific reference point in 3D space or a probability distribution over a set of reference points.
These reference points are encoded as 3D positional embeddings, which are added to the tokens and actively participate in the attention computation.

To integrate point cloud features, a natural choice is to introduce them at the same stage as image features, while using the corresponding 3D locations as reference points for positional encoding. This leads to the architecture illustrated in~\cref{fig:fusion_module}.
Through this design, we introduce an additional modality while preserving the sparse representation attribute of the Q-Former 3D block as well as its ability to leverage temporal information.
However, we observed that directly incorporating point cloud features in this way leads to training instability and convergence to inferior optima.
Inspired by LLaMA-Adapter~\cite{rZhang_2024_LLaMA-Adapter}, we therefore incorporate point cloud features through a zero-initialized gating mechanism.

Given a scene’s point cloud, image, and text prompt, we feed them into their respective encoders to obtain modality-specific representations, which are denoted as $F_P \in \mathbb{R}^{N_P \times C}$, $F_I \in \mathbb{R}^{N_I \times C}$ and $T \in \mathbb{R}^{N_T \times C}$, where $N_P$, $N_I$ and $N_T$ denotes the lengths of three modalities and $C$ equals the feature dimension of the LLM.
For \OurFusion, the inputs to the first layer consist of learnable tokens $L \in \mathbb{R}^{N_L \times C}$ and memory tokens $M \in \mathbb{R}^{N_M \times C}$, where $M$ is either inherited from the previous frame or zero-initialized when no preceding frame exists.
The learnable tokens are then updated by the first layer as follows:
\begin{align}
    L \leftarrow \operatorname{MHA}\left(\ L, [L;M], [L;M]\ \right)
\end{align}
where $\operatorname{MHA(\mathcal{Q},\mathcal{K},\mathcal{V})}$ indicates a standard multi-head attention.
When the pretraining starts with only the image modality available, the second layer of the Q-Former 3D block functions as a standard cross-attention layer and generates output as follows:
\begin{align}
    O_I &= \operatorname{MHA}\left(\ L, F_I, F_I\ \right)
\end{align}

\begin{figure*}[t] 
\centering
    \includegraphics[width=0.8\textwidth]{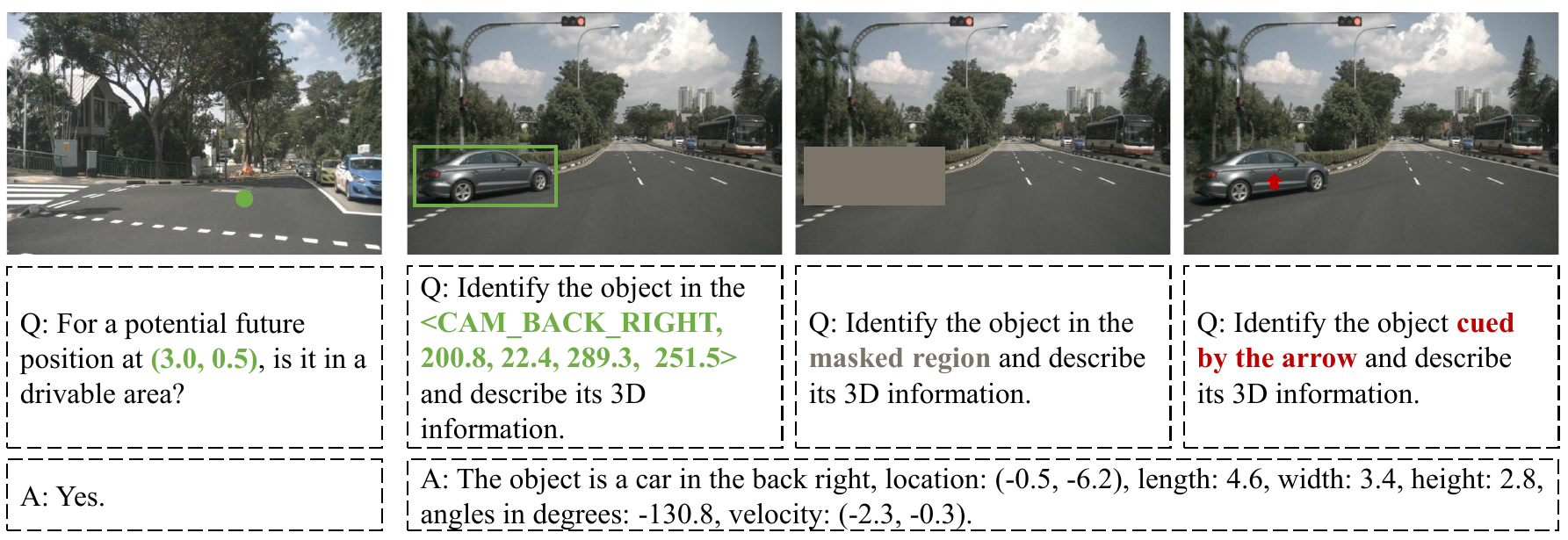}
    \\
    \makebox[\textwidth]{(a) Spatial-Perception Question-Answering}
    \\
    \includegraphics[width=0.9\textwidth]{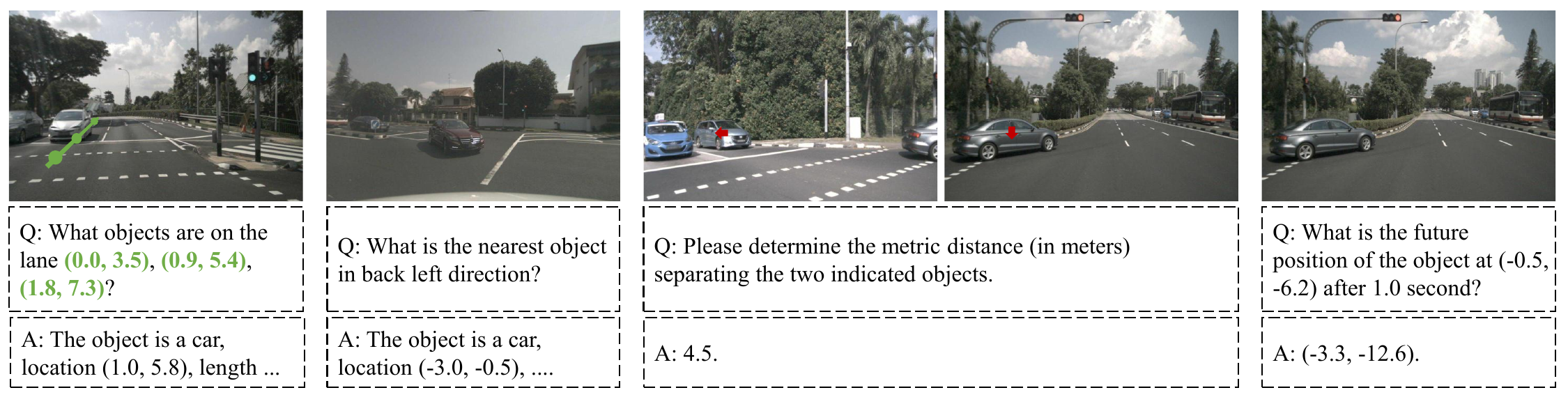}
    \\
    \makebox[\textwidth]{(b) Spatial-Reasoning Question-Answering}
    \caption{\textbf{Examples illustrating different question-answering pairs in \OurDataset.} The green dots, box, and line are highlighted only for visualization. The masked region and the red arrows are visible to the vision encoder.}
    \label{fig:dataset_examples}
    \vspace{-0.5em}
\end{figure*}

In the joint training stage, we further introduce point features as additional keys and values to the second layer, as illustrated in~\cref{fig:fusion_module}.
First, a set of linear layers is applied to transform the inputs into queries, keys and values:
\begin{align}
    \mathcal{Q} &= \operatorname{Linear_{q}}(\ L\ ),\\
    \mathcal{K}_I = \operatorname{Linear_{k}}&(\ F_I\ ), \mathcal{K}_P = \operatorname{Linear_{k}}(\ F_P\ ),\\
    \mathcal{V}_I = \operatorname{Linear_{v}}&(\ F_I\ ), \mathcal{V}_P = \operatorname{Linear_{v}}(\ F_P\ ).
\end{align}
Then, the attention scores are calculated as:
\begin{align}
    S_I &= \operatorname{softmax}(\ \mathcal{Q} \mathcal{K}_I^T/\sqrt{C}\ ), \\
    S_P &= \operatorname{softmax}(\ \mathcal{Q} \mathcal{K}_P^T/\sqrt{C}\ )
\end{align}
where $S_I, S_P \ \in \mathbb{R}^{N_L \times N_L}$.
Note that we add 3D positional embeddings derived from reference points to the queries, probability-accumulation–based 3D positional embeddings for each pixel to $\mathcal{K}_I$ and 3D positional embeddings derived from each point feature’s spatial location to $\mathcal{K}_P$.
By aligning the coordinate systems, all reference points reside in the same 3D physical space, enabling spatially aligned fusion.
For clarity, these details are omitted from the formula in this section.
The resulting scores are applied to the values to obtain the output:
\begin{align}
    O &= \operatorname{Linear_{o}}(S_I \mathcal{V}_I + S_P \mathcal{V}_P)\ \in \mathbb{R}^{N_L\times C}.
\end{align}

Because point cloud features are newly introduced and have not been jointly optimized with the image features or learnable tokens, directly incorporating them causes instability during the softmax operation.
To this end, we introduce a zero-initialized learnable gate, denoted as $g$, to adaptively control the contribution of point cloud features during training:
\begin{align}
    O &= \operatorname{Linear_{o}}\left(S_I \mathcal{V}_I + \operatorname{tanh}\left(g\right) S_P \mathcal{V}_P\right)\ \in \mathbb{R}^{N_L\times C}.
\end{align}
Since $g$ is initialized to zero, $\operatorname{tanh}(g)$ begins near zero, effectively suppressing unaligned point features at the early stage of training and enabling their gradual integration into the Q-Former. This mechanism preserves the stability of the pretrained components while allowing controlled adaptation to the new modality.
To encourage diverse and complementary feature learning, we employ independent gates for each attention head, referred to as the multi-head gate in~\cref{tab:ablation_arch}.

\subsection{\OurDataset Dataset} \label{sec:sa_nuscenes}
To enhance the spatial understanding capability of \Ours and encourage its adaptation to LiDAR point cloud data, we construct the Spatial-Aware Question-Answering dataset, abbreviated as \OurDataset.
Built upon the nuScenes dataset~\cite{nuscenes}, our \OurDataset dataset encompasses both fundamental spatial perception questions and complex spatial reasoning tasks, as illustrated in~\cref{fig:dataset_examples}.

\paragraph{Spatial Perception.}
The spatial perception question–answering tasks in \OurDataset can be categorized into two types. The first type asks whether a specific BEV point lies within a drivable area, allowing the language model to develop an awareness of road conditions. The second type requires the language model to identify an object in various ways and answer questions about its 3D attributes, including coordinates, size (length, width, and height), orientation, and velocity.
In addition, we carefully design question–answer pairs to explicitly encourage the fusion of LiDAR and image features. For example, arrows in 2D images are used to indicate objects, prompting the model to perform cross-modal reasoning between images and point clouds. Similarly, masking the corresponding region in the 2D image forces the model to extract relevant information directly from the point cloud.
Examples of spatial perception QAs are shown in~\cref{fig:dataset_examples}(a).

\paragraph{Spatial Reasoning.}
Building upon these fundamental spatial perception tasks, we further design complex spatial reasoning QA tasks to enhance the LLM’s reasoning ability, which is crucial for autonomous driving decision-making.
For instance, given a set of points along a road centerline, the model is asked to identify which objects lie on that lane.
In another type of question, the LLM must determine the nearest object in a specified direction.
We also introduce tasks involving two indicated objects (denoted by two arrows) and require the LLM to compute their 3D distance.
These objects may appear within the same image but are more often captured across different camera views, compelling the model to perform true 3D spatial reasoning rather than relying on 2D pixel distances.
Finally, we incorporate temporal reasoning tasks in which the LLM must not only identify an object but also predict its future position after a given time interval.
Examples of spatial reasoning QAs are shown in~\cref{fig:dataset_examples}(b).

\vspace{0.5em}
\noindent
All questions are generated from templates, and the answers are derived from the human-annotated ground-truth labels in the nuScenes dataset~\cite{nuscenes} and its extension OpenLane dataset~\cite{chen2022openlane}.
This design fully leverages the high-precision annotations of nuScenes, guiding the model toward spatially aware understanding.
More details on the \OurDataset dataset are provided in the supplementary material.
\section{Experiments} \label{sec:exp}
\subsection{Implementation Details} \label{sec:exp_setup}
Following the configuration of our base model OmniDrive-Agent~\cite{wang2025omnidrive}, we utilize EVA-02-L~\cite{eva02} as the vision encoder and LLaVA v1.5~\cite{hLiu_2024_LLaVA1.5} as the LLM, initializing the latter with its pretrained weights on the LLaVA-665K dataset~\cite{hLiu_2024_LLaVA1.5}.
For the point cloud encoder, we employ FSDv2~\cite{fan2024fsdv2} pretrained on nuScenes~\cite{nuscenes}.
We jointly fine-tune the entire LVLDrive—applying LoRA only to the LLM components—on the OmniDrive dataset~\cite{wang2025omnidrive} and our proposed \OurDataset dataset.
The learning rate for the \OurFusion is set to $4\times10^{-4}$, while the learning rate for image encoder, point encoder and LLM is $2\times10^{-5}$.
We use AdamW optimizer with a batch size of 16 and train for 12 epochs, using a 500-iteration warmup followed by cosine annealing policy.
Training requires approximately 30 hours on 8×A100 GPUs (40 GB each).
Additionally, we fine-tune LVLDrive on the DriveLM dataset to evaluate its performance on DriveLM.

\subsection{Dataset \& Metrics}
\paragraph{nuScenes.}
The nuScenes dataset~\cite{nuscenes} is a real-world dataset for autonomous driving research, capturing 1,000 diverse driving scenes from Boston and Singapore.
Each scene is fully annotated with 3D bounding boxes, object attributes, tracking IDs, and rich metadata, making nuScenes one of the most comprehensive datasets for 3D perception and prediction.
OpenLane~\cite{chen2022openlane} supplements the nuScenes dataset with high-quality labels for lane boundaries, centerlines, and traffic lights, which enables the generation of lane-related question-answering pairs. 
In this work, we primarily utilize the camera images and LiDAR scans from nuScenes as model inputs, and adopt the future ego-vehicle trajectories as ground truth for evaluating open-loop planning performance.

\begin{table*}
\centering
\small
\caption{
\textbf{Comparison on nuScenes open-loop planning.} $^*$ and $^{**}$ indicate not use ego status in BEV module or planner, respectively. Best results are highlighted as \colorbox{colorFst}{\bf first}, \colorbox{colorSnd}{second}, and \colorbox{colorTrd}{third}.}
\scalebox{0.98}{
\setlength{\tabcolsep}{8pt}
\begin{tabular}{l|cccc|cccc|cccc}
\toprule
\multirow{2}{*}{Method} &
\multicolumn{4}{c|}{L2 (m) $\downarrow$} & 
\multicolumn{4}{c|}{Collision Rate (\%) $\downarrow$} &
\multicolumn{4}{c}{Intersection Rate (\%) $\downarrow$} \\
                                        & 1s & 2s & 3s &\cellcolor{gray!30}Avg. & 1s & 2s & 3s& \cellcolor{gray!30}Avg. & 1s & 2s & 3s &\cellcolor{gray!30}Avg.\\
\midrule
\rowcolor{gray!30}\multicolumn{13}{l}{\textit{End-to-end Models}} \\
UniAD \cite{hu2023planninguniad}        & 0.20 & 0.42 & 0.75& 0.46 & 0.02 & 0.25 & 0.84&0.37 & \fs 0.20& \fs 1.33& \fs 3.24& \fs 1.59  \\
VAD-Base \cite{jiang2023vad}            &  0.17 &  0.34 & 0.60 &0.37 & 0.04 & 0.27 & 0.67 & 0.33 & \nd 0.21 &\rd 2.13& \nd 5.06& \nd 2.47  \\
Ego-MLP$^*$ \cite{zhai2023rethinking}   & \rd 0.15 &  0.32 & 0.59  & 0.35& \fs 0.00 & 0.27 & 0.85&0.37 & \rd 0.27& 2.52&6.60& 2.93\\
BEV-Planner~\cite{li2024bevplanner}     & 0.16 &  0.32& 0.57 & 0.35& \fs 0.00 & 0.29 & 0.73 &0.34 &0.35&2.62&6.51&3.16 \\
\midrule
\rowcolor{gray!30}\multicolumn{13}{l}{\textit{Language Models}} \\
DriveVLM~\cite{tian2024drivevlm}        & 0.18 & 0.34 & 0.68 & 0.40 & 0.10 & 0.22 & \fs 0.45 & \nd 0.27 & - & - & - & - \\
OmniReason \cite{liu2025omnireason}     & \rd 0.15 & \rd 0.31 & 0.57 & 0.34 & 0.04 & \rd 0.18 & 0.98 & 0.40 & 0.61 & 2.75 & 6.19 & 3.18 \\
Orion$^{**}$~\cite{hFu_2025_ORION}      & 0.17 & \rd 0.31 & \nd 0.55 & \nd 0.32 & 0.05 & 0.25 & 0.80 & 0.37 & - & - & - & - \\
OmniDrive-Agent~\cite{wang2025omnidrive}& \nd 0.14 & \nd 0.29 & \nd 0.55 & \rd 0.33 & \fs 0.00 & 
\fs 0.13 & \rd 0.78 & \rd 0.30 & 0.56 & 2.48 & 5.96 & 3.00 \\
\textbf{\Ours (Ours)}                   & \fs 0.13 & \fs 0.26 & \fs 0.49 & \fs 0.29 & 0.02 & \nd 0.16 & \nd 0.57 & \fs 0.25 &  0.53 & \nd 2.03 & \rd 5.22 & \rd 2.59 \\
\bottomrule
\end{tabular}
}
\label{tab:planning}
\end{table*}

\begin{table}[t]
\centering
\scriptsize
\caption{\textbf{Comparison on DriveLM dataset~\cite{cSima_2024_DriveLM}.} $^*$ denotes results reproduced by us. All other metrics are quoted directly from their original publications.}
\resizebox{\linewidth}{!}{
\begin{tabular}{l|ccc}
\toprule
Method                                          & BLEU-4 $\uparrow$ & ROUGE-L $\uparrow$ & CIDEr $\uparrow$  \\ 
\midrule
EM-VLM4AD~\cite{aGopalkrishnan_2024_EM-VLM4AD}  & 45.36 & 71.98   & 3.20  \\
MiniDrive~\cite{eZhang_2025_MiniDrive}          & 50.20 & 73.50   & 3.32  \\
MPDrive~\cite{zZhang_2025_MPDrive}              & 52.71 & \fs 76.98   & 3.56  \\
LMAD~\cite{nSong_2025_LMAD}                     & \rd 54.49 & \nd 75.72   & \rd 3.84  \\
OmniDrive-Agent$^*$~\cite{wang2025omnidrive}          & \nd 55.52 & 72.08   & \nd 15.44 \\ 
\textbf{\Ours (Ours)}                           & \fs 59.01 & \rd 73.67   & \fs 21.65 \\ 
\bottomrule
\end{tabular}
}
\label{tab:drivelm}
\end{table}

\paragraph{OmniDrive.}
The OmniDrive dataset~\cite{wang2025omnidrive} extends nuScenes by selecting planning-oriented key-frames and utilizing GPT4 and human-in-the-loop to enrich each key scene with scene description, key objects description, and reasoning over real and counterfactual trajectories.
Its annotations on trajectory plausibility—covering rule violations, collision risks, and behavioral rationality—provide richer planning-oriented supervision for training LLM-based driving models.

\paragraph{DriveLM.}
The DriveLM dataset~\cite{cSima_2024_DriveLM} formulates autonomous driving as a graph VQA task, where scene understanding and decision-making are expressed as a directed graph of QA pairs spanning perception, prediction, planning, and motion.
Built on nuScenes~\cite{nuscenes}, DriveLM provides 4871 keyframes annotated with multi-stage QA graphs grounded in 3D bounding boxes and multi-view images, enabling structured evaluation of reasoning across the full driving pipeline.
We follow the dataset split used in EM-VLM4AD \cite{aGopalkrishnan_2024_EM-VLM4AD}, employing 90\% of the data for fine-tuning and evaluating on the 5\% test set.
For other previous works, we directly cite the results reported in their papers after confirming that they were obtained under the same setting.

\paragraph{Metrics.}
For open-loop planning on nuScenes, our model outputs six future waypoints in a question–answer format, corresponding to the next 3 seconds at 0.5-second intervals.
We compute the BEV L2 distance between each predicted waypoint and the ground-truth (GT) trajectory—also known as the Displacement Error—along with the collision rate over the entire 3-second horizon.
Following BEV-Planner~\cite{li2024bevplanner}, we adopt an improved collision rate—where any collision at a given step marks all subsequent steps as collisions—and additionally evaluate the intersection rate (IR) with road boundaries, which measures whether the predicted trajectory remains within the drivable area.
In addition, to assess spatial perception, we parse bounding boxes from the VLM-generated answers responding to 3D grounding questions and compute the BEV mean Intersection over Union (mIoU), a standard metric for evaluating object detection and localization quality.
More details on this grounding evaluation are provided in the supplementary material.
On the DriveLM dataset, we evaluate how closely the generated answers match human annotations using standard language metrics, including BLEU-4~\cite{papineni2002bleu}, ROUGE-L~\cite{lin2004rouge}, and CIDEr~\cite{vedantam2015cider}.

\subsection{Open-loop Planning} \label{sec:exp_planning}
We compare \Ours with state-of-the-art end-to-end planning models and LLM-based planning approaches on the nuScenes open-loop planning benchmark in~\cref{tab:planning}.
\Ours delivers competitive results and outperforms existing LLM-based approaches.
Although our model exhibits a higher Intersection Rate than end-to-end models, LLM-based approaches offer more comprehensive functionality; for instance, our model can perform scene-level question answering.
Relative to our base model OmniDrive-Agent, LVLDrive achieves consistent gains across all average metrics, validating the effectiveness of the proposed fusion mechanism and spatial-aware fine-tuning strategy.
These findings underscore the crucial role of accurate spatial understanding and reasoning in enabling safe and robust planning.

\begin{table}[t]
\centering
\caption{\textbf{Ablation on Q-Former configuration, input modalities and gate variants.} CR and IR denote collision rate and intersection, respectively. The L2 metric for open-loop planning is consistently around 0.32 meters and is therefore omitted for space. The lighter unified Q-Former, when equipped with a zero-initialized gate, outperforms using two independent Q-Formers for the image and LiDAR modalities, demonstrating the effectiveness of \OurFusion.}
\Large
\resizebox{\linewidth}{!}{
\begin{tabular}{lllccc}
\toprule
\multirow{2}{*}{Q-Former}    & \multirow{2}{*}{Modality} & \multirow{2}{*}{\makecell{Zero-init.\\Gate}} & \multicolumn{2}{c}{Open-loop planning} & Grounding \\
\cmidrule(lr){4-5} \cmidrule(lr){6-6}
                             &     &            & CR (\%) $\downarrow$ & IR (\%) $\downarrow$ & mIoU $\uparrow$  \\
\midrule
\multirow{3}{*}{Independent} & C   & N/A        & 0.33       & 3.32          & 0.18      \\
                             & L   & N/A        & 0.39       & 3.26          & 0.17      \\
                             & C+L & N/A        & \nd 0.29       & \rd 3.20          & \rd 0.19      \\
\midrule
\multirow{3}{*}{\makecell{\textbf{Unified}\\ \textbf{(Ours)}}} & \multirow{3}{*}{C+L} & No & 0.40       & 3.45          & 0.17      \\
                             &     & Single     & \rd 0.30       & \nd 2.99          & \fs 0.21      \\
                             &     & Multi-head & \fs 0.28       & \fs 2.93          & \fs 0.21      \\ 
\bottomrule
\end{tabular}
}
\label{tab:ablation_arch}
\end{table}

\subsection{Results on DriveLM Dataset} \label{sec:exp_drivelm}

\Cref{tab:drivelm} shows that \Ours achieves competitive performance on BLEU-4 and ROUGE-L, indicating that its generated answers are comparable to prior vision–language driving models in terms of surface-level phrasing and syntactic overlap.
Notably, \Ours attains a CIDEr score of 21.65, substantially surpassing all baselines. As CIDEr places greater emphasis on content relevance, informativeness, and agreement with human-consensus keywords, this significant margin suggests that LVLDrive generates answers that more accurately capture critical spatial cues, object attributes, and driving-relevant semantics. In particular, the improvement reflects stronger grounding in 3D geometry and more precise references to spatial relationships that are essential for safe driving.
Overall, while lexical similarity remains comparable across models, \Ours demonstrates superior 3D scene understanding and spatial reasoning. These results validate the effectiveness of our LiDAR injection strategy and spatial-aware fine-tuning in enhancing semantically faithful and geometrically grounded language generation.

\subsection{Ablation Study} \label{sec:ablation}

\paragraph{Ablation on Model Configuration.}
In~\cref{tab:ablation_arch}, we examine how different input modalities, Q-Former configuration and gating variants affect both open-loop planning and 3D grounding.
The first three rows compare models using camera input only (C), LiDAR input only (L), and their combination (C+L), where image and LiDAR features are retrieved by separate Q-Formers.
Adding LiDAR as an additional modality yields moderate improvements, demonstrating that it provides complementary geometric cues beneficial for both planning and grounding.
However, the additional Q-Former also brings extra computation and memory consumption, making it a suboptimal solution.
In the unified Q-Former setting (bottom three rows), we further investigate the impact of different gating mechanisms.
The multi-head gate achieves the best overall performance, reducing collision rate from 0.40\% to 0.28\% and intersection rate from 3.45\% to 2.93\% compared to the no-gate baseline, while improving mIoU from 0.17 to 0.21.
These results indicate that the proposed zero-initialized multi-head gate improves training stability, leading to more robust planning and grounding.
Additionally, the lightweight unified Q-Former, when equipped with a zero-initialized gate, surpasses the separate Q-Former design, further demonstrating the effectiveness of \OurFusion.

\begin{table}[t]
\centering
\caption{\textbf{Ablation on dataset composition.} P. and R. stand for perception and reasoning QAs, respectively. CR and IR denote collision rate and intersection, respectively. The L2 metric for open-loop planning is consistently around 0.32 meters and is therefore omitted for space.}
\resizebox{\linewidth}{!}{
\begin{tabular}{ccccccc}
\toprule
\multirow{2}{*}{OmniDrive} & \multicolumn{2}{c}{\OurDataset} & \multicolumn{2}{c}{Open-loop planning} & Grounding \\
\cmidrule(lr){2-3} \cmidrule(lr){4-5} \cmidrule(lr){6-6}
        & P.     & R.  & CR (\%) $\downarrow$ & IR (\%) $\downarrow$ & mIoU $\uparrow$  \\
\midrule
\cmark  &           &          & 0.39       & 3.17          & 0.00      \\
\cmark  & \cmark    &          & \nd 0.29       & \rd 2.89          & \nd 0.20       \\
\cmark  &           & \cmark   & \rd 0.33       & \nd 2.84          & \rd 0.11      \\
\cmark  & \cmark    & \cmark   & \fs 0.27       & \fs 2.83          & \fs 0.22      \\
\bottomrule
\end{tabular}
}
\label{tab:ablation_data}
\vspace{-0.5em}
\end{table}

\paragraph{Ablation on Dataset Composition.}
\Cref{tab:ablation_data} evaluates the contributions of different components in our \OurDataset dataset by separating perception-oriented QAs (\eg 3D grounding, drivable-area judgment) from reasoning-oriented QAs (\eg estimating inter-object distances, predicting agent intentions).
Interestingly, when trained solely on the OmniDrive dataset~\cite{wang2025omnidrive}, \Ours collapses on the grounding task.
Although the model can still produce seemingly reasonable textual responses, it consistently omits key information required to reconstruct a 3D box—such as size or orientation—making it impossible to parse a valid box from the output.
As a result, the grounding metric drops to zero.
Incorporating perception QAs notably improves all metrics, reducing the collision rate from 0.39\% to 0.29\% and the intersection rate from 3.17\% to 2.89\%.
Incorporating reasoning QAs further boosts overall performance relative to training solely on OmniDrive. Nevertheless, compared with adding only perception QAs, the CR and mIoU are still inferior. We attribute this to the scarcity of reasoning questions that demand accurate 3D information about individual objects, which limits the model’s ability to learn robust instruction-following behavior.
When both perception and reasoning QAs are used during training, the model achieves the best overall results, with the lowest CR (0.27\%), lowest IR (2.83\%), and highest mIoU (0.22).
These findings demonstrate that perception and reasoning QAs offer complementary supervision signals that jointly enhance spatial understanding and decision-making.
\section{Conclusion} \label{sec:conclusion}

In this work, we introduced \Ours, a LiDAR-Vision-Language framework that enhances Vision-Language Models (VLMs) with robust 3D metric spatial understanding for autonomous driving. By integrating LiDAR information through our~\OurFusion, \Ours effectively fuses LiDAR and image inputs while preserving the linguistic and visual reasoning capabilities of the pre-trained VLM backbone. In addition, we proposed a spatial-aware question-answering (SA-QA) dataset that explicitly guides the model to interpret and utilize LiDAR point clouds through carefully designed QA pairs covering 3D grounding, spatial reasoning, and future state prediction. Our experiments demonstrate the effectiveness of integrating explicit 3D spatial information into VLMs, leading to improved reasoning and decision-making in autonomous driving systems.

\section*{Acknowledgements}

{
    \small
    \bibliographystyle{ieeenat_fullname}
    \bibliography{main}
}

\clearpage
\setcounter{page}{1}
\maketitlesupplementary
\renewcommand{\thefigure}{S\arabic{figure}}
\renewcommand{\thetable}{S\arabic{table}}
\renewcommand{\thesection}{S\arabic{section}}

\section{Abstract} \label{sec:supp_abstract}
This supplementary material provides additional details and analyses of our work.
Further information about our proposed \OurDataset dataset is presented in~\cref{sec:supp_our_dataset}, and a detailed comparison with NuScenes-SpatialQA is provided in~\cref{sec:supp_comp_spatialqa}.
Additional details on the 3D grounding benchmark are included in~\cref{sec:supp_grounding_benchmark}, and limitations and future work are discussed in~\cref{sec:limitations}.

\begin{table*}[t]
\centering
\footnotesize
\caption{\textbf{Generation logic for spatial perception (SP) tasks in the SA-QA dataset.}}
\begin{tabularx}{\textwidth}{@{}l 
    >{\raggedright\arraybackslash\hsize=0.9\hsize}X 
    >{\raggedright\arraybackslash\hsize=0.6\hsize}X 
    >{\raggedright\arraybackslash\hsize=1.6\hsize}X 
    >{\raggedright\arraybackslash\hsize=0.9\hsize}X@{}}
\toprule
\textbf{ID} & \textbf{Prompt Template} & \textbf{Input Data Transformation} & \textbf{Generation Logic} & \textbf{Answer Generation} \\ \midrule

\textbf{SP-01} & 
``For a potential future position at $(x, y)$, is it in a drivable area?'' & 
None. & 
1. Randomly sample a query point in the front region of the ego-vehicle ($5{\le}X{\le}20$, $-5{\le}Y{\le}5$). \newline
2. Generate a drivable mask in BEV by buffering lane centerlines with a $1.75\text{m}$ margin. \newline
3. Check point inclusion against the mask. & 
\textbf{Binary:} ``Yes'' if inside; ``No'' otherwise. \\ \addlinespace

\textbf{SP-02} & 
``Identify the object in $<$\textit{CAM}, $x_{min}, y_{min}, x_{max}, y_{max}$$>$ and describe its 3D information.'' & 
None. & 
1. Project 3D annotations to the 2D planes of all 6 cameras. \newline
2. Compute 2D bounding boxes $[x_{min}, y_{min}, x_{max}, y_{max}]$ clamped to image dims, filtering out candidates that are invisible or too small. \newline
3. Sample an object and format the answer. \newline
4. Format the prompt with the target object's coordinates. & 
\textbf{Text:} ``The object is a $<$category$>$ in the $<$CAM$>$, location: $(x,y)$, length: $<$$l$$>$, width: $<$$w$$>$, height: $<$$h$$>$, angles in degree: $<$$yaw$$>$.'' (Rounded to 0.1). \\ \addlinespace

\textbf{SP-03} & 
``Identify the object cued by the \textbf{arrow} and describe its 3D information.'' & 
Draw an arrow on the image. & 
1--3. Follow steps 1--3 of SP-02 to filter candidates and sample one object. \newline
4. Draw a visual arrow on the image pointing to the center of the target's 2D mask. \newline
5. Construct the prompt referencing the arrow cue. & 
\textbf{Text:} Same format as SP-02. \\ \addlinespace

\textbf{SP-04} & 
``Identify the object in the \textbf{masked region} and describe its 3D information.'' & 
Mask a region of the image. & 
1--3. Follow steps 1--3 of SP-02 to filter candidates and sample one object. \newline
4. Apply a mask to the target's 2D bounding box region (forcing LiDAR reliance). \newline
5. Construct the prompt referencing the masked region. & 
\textbf{Text:} Same format as SP-02. \\ 
\bottomrule
\end{tabularx}
\label{tab:saqa_perception}
\vspace{1em}
\end{table*}

\begin{table*}[t]
\centering
\footnotesize
\caption{\textbf{Generation logic for spatial reasoning (SR) tasks in the SA-QA dataset.}}
\begin{tabularx}{\textwidth}{@{}l 
    >{\raggedright\arraybackslash\hsize=0.9\hsize}X 
    >{\raggedright\arraybackslash\hsize=0.6\hsize}X 
    >{\raggedright\arraybackslash\hsize=1.6\hsize}X 
    >{\raggedright\arraybackslash\hsize=0.9\hsize}X@{}}
\toprule
\textbf{ID} & \textbf{Prompt Template} & \textbf{Input Data Transformation} & \textbf{Generation Logic} & \textbf{Answer Generation} \\ \midrule

\textbf{SR-01} & 
``What objects are on the lane defined by points $(x_1, y_1), (x_2, y_2), (x_3, y_3)$?'' & 
None. & 
1. Randomly select a lane centerline from the OpenLane~\cite{chen2022openlane} map annotations. \newline
2. Aggregate the associated objects located on the selected lane centerline from the OpenLane annotation set~\cite{chen2022openlane}. \newline
3. Format the prompt and answer. & 
\textbf{List:} ``The object is a $<$category$>$, location...''. \\ \addlinespace

\textbf{SR-02} & 
``What is the nearest object in the $<$DIRECTION$>$ direction?'' & 
None. & 
1. Define 4 spatial sectors (\eg Front-Left, Back-Left) relative to the current ego-vehicle heading. \newline
2. Filter objects located within the target area. \newline
3. Calculate Euclidean distances for all candidates and sort in ascending order. \newline
4. Format the prompt and answer. & 
\textbf{Text:} Description of the object with index 0 (minimum distance), using the format from SP-02. \\ \addlinespace

\textbf{SR-03} & 
``Please determine the metric distance (in meters) separating the two indicated objects.'' & 
Draw arrows on the images. & 
1. Select two distinct visible objects ($O_A, O_B$), potentially across different camera views. \newline
2. Draw visual arrows pointing to $O_A$ and $O_B$ in their respective images (following SP-03). \newline
3. Compute the L2 norm between their 3D centroids: $\| C_A - C_B \|_2$. & 
\textbf{Scalar:} ``$D$.'' (The value is rounded to 0.1 meters). \\ \addlinespace

\textbf{SR-04} & 
``What is the future position of the object at $(x, y)$ after $T$ second?'' & 
None. & 
1. Randomly sample one object that possesses a future trajectory within the nuScenes annotation set \cite{nuscenes}. \newline
2. Randomly select a future position and time interval for the sampled object, and subsequently structure the input prompt and the corresponding target answer. &
\textbf{Coordinate:} ``$(x_{fut}, y_{fut})$.'' \\ 
\bottomrule
\end{tabularx}
\label{tab:saqa_reasoning}
\vspace{1em}
\end{table*}
\section{More Details on \OurDataset Dataset}
\label{sec:supp_our_dataset}

This dataset is built on top of nuScenes~\cite{nuscenes} dataset and enriched using the ground-truth annotations provided by both nuScenes and OpenLane~\cite{chen2022openlane} to generate question-answering (QA) pairs.
As SA-QA is designed for training, we use only the training split of the nuScenes dataset.
The specific QA formats and the step-by-step generation procedure are summarized in~\cref{tab:saqa_perception} and~\cref{tab:saqa_reasoning}.
Note that before generating any QA pairs, we convert all annotations and LiDAR point clouds into the ego coordinate system.

\section{Comparison with NuScenes-SpatialQA}
\label{sec:supp_comp_spatialqa}
NuScenes-SpatialQA \cite{kTian_2025_NuScenes-SpatialQA} is a concurrent study that shares certain similarities with our proposed SA-QA dataset.
While both datasets focus on the spatial reasoning of Vision-Language Models (VLMs), \OurDataset diverges by prioritizing instruction tuning, explicit metric grounding, and cross-modal alignment.

\paragraph{Training vs. Evaluation Focus.} 
NuScenes-SpatialQA is constructed exclusively on the nuScenes validation split (150 scenes).
Its primary purpose is to serve as a zero-shot benchmark to evaluate existing general-purpose VLMs.
In contrast, \OurDataset is designed as a massive-scale instruction-tuning dataset constructed on the training split (850 scenes).
This scale allows \Ours to learn complex spatial relationships rather than merely being tested on them.

\paragraph{Enhanced Cross-Modality Interaction.}
To promote robust alignment across modalities, SA-QA introduces specific task designs targeting Text-Vision-LiDAR integration.
We implement modality masking (\cref{tab:saqa_perception}, SP-04), which masks the target in the image to encourage the model to retrieve geometric information directly from the LiDAR data.
Furthermore, we utilize visual cues (\cref{tab:saqa_perception}, SP-03; \cref{tab:saqa_reasoning}, SR-03), where prompts explicitly reference arrows drawn on the image.
This design drives the model to bind language with joint image–LiDAR features, enabling richer 2D–3D spatial reasoning.

\paragraph{Explicit 3D Grounding vs. Relative Depth.}
While NuScenes-SpatialQA utilizes 3D annotations to generate answers, its questions are largely limited to relative 3D distances or topological relationships (\eg ``Is object A closer than object B?"), omitting explicit 3D locations.
This allows models to rely on approximate depth cues without mastering metric space.
Conversely, \OurDataset dataset requires explicit 3D grounding, asking the model to output precise coordinates $(x,y)$ and absolute dimensions (\cref{tab:saqa_perception}, SP-02).
This forces the model to internalize a true metric understanding of the 3D environment.

\paragraph{Global Perception vs. Intra-View Limitations.}
A critical limitation of NuScenes-SpatialQA is that questions are typically restricted to the same camera view.
This relies on local visual comparisons and omits a holistic perception of the surrounding environment.
\OurDataset explicitly challenges this by constructing cross-view reasoning tasks (\cref{tab:saqa_reasoning}, SR-03) where target objects may appear in disparate sensors (\eg Front Camera vs. Back-Right Camera).
To answer these correctly, the model cannot rely on a single 2D image but must fuse information into a unified global coordinate system.

\paragraph{Automated Efficiency vs. LLM Latency.}
NuScenes-SpatialQA relies on heavy Large Language Models to generate dense captions and formulate questions, a process that introduces significant computational latency and cost.
In contrast, \OurDataset is fully automated and rule-based.
By deriving QA pairs directly from ground-truth annotations, our generation process is computationally negligible.
This high efficiency supports dynamic augmentation, allowing us to apply random sampling strategies on the fly—specifically randomizing target objects, lane segments, and temporal intervals for future prediction.

\section{More Details on 3D Grounding Benchmark}
\label{sec:supp_grounding_benchmark}
Throughout model development, we observed that planning-only evaluation is insufficient and fails to capture a model’s understanding of the spatial distribution of surrounding objects. To better assess this capability, we construct a grounding benchmark using the ground-truth annotations of the nuScenes~\cite{nuscenes} validation set.

\paragraph{Distance-Based Object Sampling.}
Traversing all annotated objects would lead to prohibitive inference time, so we sample target objects based on distance.
For each frame, we compute the distance from the bottom center of every annotated 3D bounding box to the ego vehicle and sort all objects by distance. The objects are then grouped into four ranges. After experimenting with interval sizes of 15 m, 20 m, and 25 m, we found that a 15 m step yields the most balanced distribution across groups, and thus adopt 15 m as the interval. Finally, for each frame, we randomly sample one object from each distance range to form our grounding benchmark. The number of objects in each range is shown in~\cref{tab:supp_grounding}.

\paragraph{Question and Answer Generation.}
For each selected object, we format its 3D bounding box parameters using the following answer template: The object is a $<$class$>$ in the $<$direction$>$, location: $(X, Y)$, length: $L$, width: $W$, height: $H$, angles in degrees: $\theta$. 
Meanwhile, we project the eight corners of the 3D box onto the image plane and compute the corresponding 2D bounding box $\left ( x_{min}, y_{min}, x_{max}, y_{max} \right )$, which is incorporated into the question. If any of these coordinates fall outside the image bounds, they are clipped to the valid image range.

\paragraph{Answer Parsing and Evaluation.}
Because similar QA patterns are also used during training, the model typically produces answers that closely follow the desired template. This allows us to reliably parse a BEV bounding box $(X,Y,L,W,\theta)$ from the predicted answer and compute the BEV mIoU.
Concretely, we first compute the IoU for each individual object. We then average the IoUs within each of the four distance ranges, and finally report the mean of these four group-wise IoUs as the overall grounding metric.

\begin{table}[t]
\centering
\scriptsize
\caption{\textbf{Distribution of object counts across distance ranges in our grounding benchmark.} }
\resizebox{\linewidth}{!}{
\begin{tabular}{ccccc}
\toprule
0-15m     & 15-30m     & 30-45m     & 45m-Inf    & Overall \\
\midrule
5,304      & 5,684       & 5,367       & 3,903       & 20,258 \\
\bottomrule
\end{tabular}
}
\label{tab:supp_grounding}
\end{table}

\section{Limitations and Future Work.} \label{sec:limitations}
Despite the benefits of our gradual fusion strategy, \Ours is still constrained by the limited availability of large-scale, naturally paired text–LiDAR data. In an ideal setting, one would pretrain a unified model on large-scale aligned language–LiDAR samples, enabling the representation space of 3D geometry and linguistic concepts to be co-optimized from scratch. The construction of such datasets remains extremely challenging, and closing this gap is an important direction for future work.

\end{document}